%% file: main.tex
\documentclass[10pt,twocolumn,letterpaper]{article}

\usepackage{iccv}
\usepackage{times}
\usepackage{epsfig}
\usepackage{graphicx}
\usepackage{amsmath}
\usepackage{amssymb}
\usepackage{amsmath,graphicx}
\usepackage{color,soul}
\usepackage{siunitx}
\usepackage{subcaption}
\usepackage{amsmath}
\usepackage{bbm}
\usepackage{subcaption}
\usepackage{algorithm}
\usepackage{algpseudocode}
\usepackage{multirow}
\usepackage{graphicx}
\usepackage{booktabs}
\usepackage{enumitem}

\usepackage[table]{xcolor}

\newcommand{\cgr}{\cellcolor[gray]{0.9}}
\newcommand\blfootnote[1]{%
  \begingroup
  \renewcommand\thefootnote{}\footnote{#1}%
  \addtocounter{footnote}{-1}%
  \endgroup
}

\usepackage[pagebackref=true,breaklinks=true,letterpaper=true,colorlinks,bookmarks=false]{hyperref}

\iccvfinalcopy 

\ificcvfinal\fi

\begin{document}

\title{Can Unstructured Pruning Reduce the Depth in Deep Neural Networks?}

\author{Zhu Liao \qquad Victor Qu\'etu \qquad Van-Tam Nguyen \qquad Enzo Tartaglione\\
LTCI, T\'el\'ecom Paris, Institut Polytechnique de Paris, France\\
{\tt\small \{name.surname\}@telecom-paris.fr}
}

\maketitle
\ificcvfinal\thispagestyle{empty}\fi

\begin{abstract}
    Pruning is a widely used technique for reducing the size of deep neural networks while maintaining their performance. However, such a technique, despite being able to massively compress deep models, is hardly able to remove entire layers from a model (even when structured): is this an addressable task? In this study, we introduce EGP, an innovative Entropy Guided Pruning algorithm aimed at reducing the size of deep neural networks while preserving their performance. The key focus of EGP is to prioritize pruning connections in layers with low entropy, ultimately leading to their complete removal. Through extensive experiments conducted on popular models like ResNet-18 and Swin-T, our findings demonstrate that EGP effectively compresses deep neural networks while maintaining competitive performance levels. Our results not only shed light on the underlying mechanism behind the advantages of unstructured pruning, but also pave the way for further investigations into the intricate relationship between entropy, pruning techniques, and deep learning performance. The EGP algorithm and its insights hold great promise for advancing the field of network compression and optimization. 
    The source code for EGP is 
\href{https://github.com/ZhuLIAO001/Unstructured_Relu_Pruning_Reduce_Depth.git}{released open-source}.\blfootnote{© 2023 IEEE. Personal use of this material is permitted. Permission from IEEE must be obtained for all other uses, in any current or future media, including reprinting/republishing this material for advertising or promotional purposes, creating new collective works, for resale or redistribution to servers or lists, or reuse of any copyrighted component of this work in other works.\\~\\This paper has been accepted for publication at the Workshop on Resource Efficient Deep Learning for Computer Vision (ICCVW).}
\end{abstract}

\input{sections/1_intro.tex}

\input{sections/3_method.tex}

\input{sections/4_results.tex}

\input{sections/5_conclusion.tex}

 \textbf{Acknowledgements.} The research leading to these results has received funding from the project titled ``PC6-FITNESS'' in the frame of the program ``PEPR 5G et Réseaux du futur''. Zhu Liao is funded by China Scholarship Council (CSC).

{\small
\bibliographystyle{ieee_fullname}
\bibliography{main}
}

\end{document}

%% file: sections/1_intro.tex
\section{Introduction}
\label{sec:intro}

Deep neural networks have become a cornerstone of artificial intelligence and machine learning, achieving remarkable success in a wide range of applications, from video recognition~\cite{beye2022recognition} and natural language processing~\cite{yang2022transformer} to robotics~\cite{renzulli2022towards} and autonomous driving~\cite{bogdoll2022anomaly}. However, the large size of these models poses significant challenges in terms of storage, memory, and computational requirements. To address these issues, researchers have developed various techniques to reduce the size of neural networks without sacrificing their performance, including quantization~\cite{gholami2022survey,yang2019quantization}, compression~\cite{deng2020model,tartaglione2021hemp}, and pruning~\cite{han2015learning,hassibi1993second,lecun1990optimal,li2016pruning,tartaglione2022loss}.
\begin{figure}[t]
    \centering
    \includegraphics[width=\columnwidth]{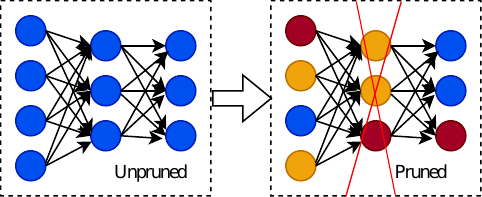}
    \caption{Effect of unstructured pruning in ReLU-activated models: some neurons are always OFF (yellow) and others are consistently in the linear region (red). These can be removed with no impact on the performance. Some neurons are in both regions (blue). If no neuron is in both regions, the layer can be fused with the next one (it is a linear combination), and the depth can be decreased (the entire layer is removed - red cross).}
    \label{fig:teaser}
\end{figure}

Pruning, in particular, has emerged as a popular and effective method for reducing the number of parameters in a neural network~\cite{bragagnolo2021role,lecun1990optimal,molchanov2017variational}. The idea behind pruning is to selectively remove weights, neurons, or entire layers based on their magnitude or other criteria~\cite{fan2022real,li2022revisiting,yu2022width}. The pruned model can then be fine-tuned to recover its original performance, resulting in a smaller and more efficient network. Pruning is particularly effective for deep neural networks, which tend to be over-parameterized and therefore have many redundant or irrelevant weights.

Despite the widespread use of pruning, its underlying mechanism and the factors that contribute to its benefits are not well understood. Previous work has investigated the effects of pruning on the sparsity and connectivity patterns of neural networks \cite{chen2021lottery,chen2020lottery,frankle2018the}, as well as its impact on the generalization performance of the pruned models \cite{han2015learning}. Nowadays, indeed, massive study around this subject is being conducted~\cite{laurent2023packed,tartaglione2022rise}. Although some works used entropy to drive pruning mechanisms~\cite{hur2019entropy}, to the best of our knowledge, no previous work has explored the relationship between pruning and the entropy of activations in the network. 

In this work, we investigate the impact of unstructured pruning on the entropy of activations in the network. In the context of neural networks, entropy can provide insights into the complexity and diversity of the representations learned by the model. Leveraging this measure, we propose EGP, a strategy to drive the unstructured pruning mechanism to prune layers exhibiting low entropy. Our experiments show that unstructured pruning reduces the entropy of activations in the network (Fig.~\ref{fig:teaser}). This reduction in entropy suggests that pruning results in a more structured activation pattern, which may as well improve the ability of the network to generalize to new data. Interestingly, this approach overcomes the limits of structured pruning, which removes neurons persistently OFF. 

Our work provides a new perspective on the mechanism behind the benefits of unstructured pruning and sheds light on the relationship between entropy, pruning, and deep learning performance. These findings open up avenues for further exploration. We have also developed an entropy-based iterative pruning technology that shows promise in compressing neural networks more effectively.

Despite other works proposing entropy-based approaches to drive pruning~\cite{hur2019entropy,luo2017entropybased,min20182pfpce}, to the best of our knowledge, EGP is the very first entropy-driven approach that can effectively reduce the depth of a deep neural network model without harming its performance.
Different from previous entropy-based pruning methods which tried to remove a certain amount of parameters or neurons inside layers, EGP tries to remove whole layers. EGP is tested on popular models for computer vision, like ResNet-18 and Swin-T, and opens the roads to a new class of pruning algorithms.

%% file: sections/3_method.tex
\section{Entropy Guided Pruning}
\label{sec:method}

\begin{figure}
    \centering
    \begin{subfigure}{\columnwidth}
    \includegraphics[width=\textwidth]{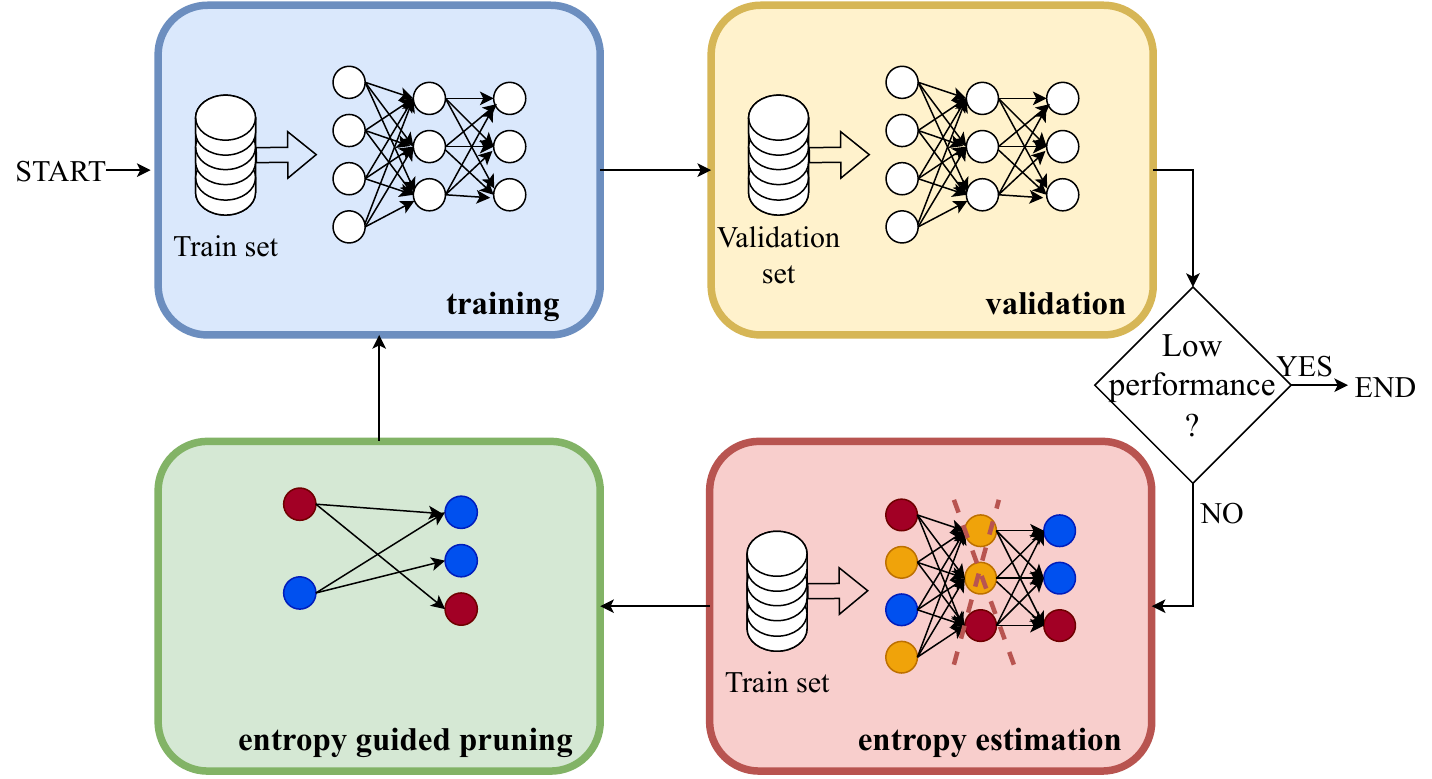}
    \end{subfigure}
    \caption{Overview of the EGP approach.}
    \label{fig:Workflow}
\end{figure}

In this section, we will provide details on how we compute the entropy of the activations inside the neural network model and develop our entropy-based iterative pruning. Fig.~\ref{fig:Workflow} proposes an overview of the proposed entropy computing and pruning method. First, we introduce the entropy calculation method (Sec.~\ref{sec:setup}), showing also how to handle critical cases (Sec.~\ref{sec:EntropyGoesZero}), and then we provide an overview of the entropy-based iterative pruning method (Sec.~\ref{sec:EntropyIterativePruning}).

\subsection{Derivation}
\label{sec:setup}
Let us extract the output $\boldsymbol{y}_{l,i}^\xi$ of the $i$-th neuron in the $l$-th layer (where $N_l$ is the number of neurons in the $l$-th layer), given as input the $\xi$-th sample. Let us assume the activation function $\phi_n(\cdot)$ for the $l$-th layer is ReLU. Toward this end, we can write
\begin{equation}
    \boldsymbol{y}_{l,i}^\xi = \text{ReLU}\left(\boldsymbol{z}_{l,i}^\xi\right),
\end{equation}
where $\boldsymbol{z}_{l,i}$ is the \emph{post-synaptic potential} for the $i$-th neuron in the $l$-th layer. Please note that in convolutional layers the dimensionality for $\boldsymbol{y}_{l,i}^\xi$ is not necessarily unitary, but is proportional to the input's size $M_l=K_{1,l}\times K_{2,l}$, where $K_{1,l}$ and $K_{2,l}$ are the dimensions for the generated feature map. We assume, for the sake of simplicity, that the size of every $\xi$ is the same for the whole dataset $\Xi$. We know that ReLU-activated neurons have essentially two working regions:
\begin{itemize}[noitemsep,nolistsep]
    \item an ON region, for $z_{l,i,j}> 0$;
    \item an OFF region, for $z_{l,i,j} \leq 0$.
\end{itemize}
These working regions will be our possible ``states''. To identify the state $s_{l,i,j}$ for the $j$-th feature extracted by the $i$-th neuron, we can apply the one-step function $H(\cdot)$ to the post-synaptic potential, obtaining $s_{l,i,j}^\xi = H(z_{l,i,j}^\xi)$. At this point, we can obtain the frequency of the $i$-th neuron to be in the ON state, simply through
\begin{equation}
    \label{eq:probsymb}
    p^{\text{ON}}(l,i| \Xi) = \frac{1}{\|\Xi\|_0\cdot M_l}\sum_{\xi\in \Xi} \sum_j (s_{l,i,j}^{\xi}),
\end{equation}
where $\|\Xi\|_0$ is the cardinality of the dataset. Evidently, we know that in this case $p^{\text{OFF}}(l,i| \Xi) = 1 - p^{\text{ON}}(l,i| \Xi)$. At this point, we can use the definition in \eqref{eq:probsymb} to write the entropy $\mathcal{H}(l,i|\Xi)$:
\begin{align}
    \label{eq:entr}
    \mathcal{H}(l,i|\Xi) =& 
    -\frac{1}{\log(2)} \left\{p^{\text{ON}}(l,i|\Xi) \log\left[p^{\text{ON}}(l,i|\Xi)\right ] +\right .\nonumber\\
    &\left .+ p^{\text{OFF}}(l,i|\Xi) \log\left [p^{\text{OFF}}(l,i|\Xi)\right ]\right\} .
\end{align}

\subsection{When the entropy goes to zero}
\label{sec:EntropyGoesZero}
The evaluation of the entropy proposed in \eqref{eq:entr} is not intended as a measure of information flowing through the given layer, but it gives us an important indication of the effective use of the ReLU non-linearity. More specifically, for the $i$-th neuron in the $l$-th layer, we will have $\mathcal{H}(l,i|\Xi)=0$ in two possible cases:
\begin{itemize}[noitemsep,nolistsep]
\item $s_{l,i,j}^\xi = 0, \forall j,\xi$. This case is achieved in three possible ways: either all the parameters $\boldsymbol{w}_{l,i}$ of the $i$-th neuron are zero, or the input $\boldsymbol{y}_{l-1}^\xi$ is zero $\forall \xi$, or again $z_{l,i,j}^\xi\leq 0, \forall j,\xi$. In these cases, the neuron's output will always be zero, and the entire neuron can be removed/pruned from the model, with no performance loss.\\
\item $s_{l,i,j}^\xi = 1,\forall j,\xi$. This case is achieved if $z_{l,i,j}^\xi>0, \forall j,\xi$. In this state, the $i$-th neuron in the $l$-th layer uses only the linear region, becoming a linear neuron: its contribution can be ``absorbed'' by the next layer (as the neuron and the next layer are a linear combination), evidencing a collapse in the neural network's depth.
\end{itemize}
In this work we will be looking for both: while the first is expected, and known to rise in high pruning regimes~\cite{bragagnolo2021role,tartaglione2022loss}, the second one is more surprising, but can potentially lead to similar gains as those obtained with pruning.\\
It is possible to extend the proposed formulation to other activations, like the sigmoid and GeLU (we will test on Swin-T in Sec.~\ref{sec:res}).



\subsection{Entropy-based iterative pruning}
\label{sec:EntropyIterativePruning}
As our target is to reduce the depth of deep neural networks by removing zero-entropy layers, we implement iterative pruning based on the entropy of different layers to get more layers with entropy zero. 

Let us set the percentage of parameters to be pruned at each pruning iteration as $\zeta$, and the total weight parameters of the considered $L$ layers in the model as $\|\theta\|_0$. Then, in each pruning iteration, the number of weight parameters to be pruned $\|\theta\|_0^{\text{pruned}}$ is given by
\begin{equation}
    \label{eq:totalPrunParas}
    \|\theta\|_0^{\text{pruned}} = \zeta \cdot \|\theta\|_0 .
\end{equation}
After calculating the entropy of each neuron belonging to the $l$-th layer, we can calculate an average entropy for it, simply through
\begin{equation}
    \hat{\mathcal{H}}(l|\Xi) = \frac{1}{N_l}\sum_i \mathcal{H}(l,i|\Xi).
\end{equation}

We would like to route the magnitude pruning algorithm towards removing more parameters in layers where the entropy is low. This is because manifesting a low entropy is a symptom of not necessarily requiring a non-linear activation in these layers, and for such reason, these layers are the most promising to remove. However, we know that $\mathcal{H}(l,i|\Xi)\geq 0 \forall l, i, \Xi$: this means that, when having $\hat{\mathcal{H}}(l|\Xi) = 0 \Leftrightarrow \mathcal{H}(l,i|\Xi) = 0 \forall i$. When this happens, we know that we are able to completely remove such a layer, and we no longer need to prune it.

In order to increase the number of layers with zero entropy, more pruning should be applied to layers with lower entropy (more likely to reach zero entropy). Simultaneously, to minimize the impact on model performance, more pruning should be done on smaller magnitude weights. To achieve both goals, we propose a \emph{pruning irrelevance} meter $\mathcal{I}_l$ for the $l$-th layer 
\begin{equation}
    \label{eq:EntroMag}
    \mathcal{I}_{l} = \hat{\mathcal{H}}(l|\Xi) \cdot  \frac{1}{\|\theta_l\|_0}\sum_{i} |\theta_{l,i}|,
\end{equation}
where $\|\theta_l\|_0$ is the cardinality of the non-zero weights in the $l$-th layer. Effectively, the larger this value is, the least we are interested in removing parameters from it. However, we are very interested in removing parameters from layers having very low pruning irrelevance: for such reason, we define the complementary measure of \emph{pruning relevance}
\begin{equation}
    \label{eq:EntroMag_hat}
    \mathcal{R}_l = \left\{
    \begin{array}{c l}
        \frac{\sum_j \mathcal{I}_j}{\mathcal{I}_l} & \mathcal{I}_l\neq 0\\
        0   & \mathcal{I}_l = 0 .
    \end{array}
    \right .
\end{equation}
Finally, in order to assess the exact amount of parameters to be removed at the $l$-th layer we resort to a softmax smoothening, according to
\begin{equation}
    \label{eq:LayerPrunParas}
     \|\theta_l\|_0^{\text{pruned}} =  \|\theta\|_0^{\text{pruned}} \cdot  \frac{\exp[\mathcal{R}_l-\max_k(\mathcal{R}_k)]}{\sum_{j}\exp[\mathcal{R}_j-\max_k(\mathcal{R}_k)]} .
\end{equation}
If the number of parameters pruned assigned to a specific layer exceeds its remaining parameter count, then prune all parameters of that layer. Afterward, distribute the remaining number of pruning parameters to other layers following the methods outlined in~\eqref{eq:EntroMag} and~\eqref{eq:LayerPrunParas}.

%% file: sections/4_results.tex
\section{Experiments}
\label{sec:res}

\begin{figure}[t]
\centering
\includegraphics[width=\columnwidth]{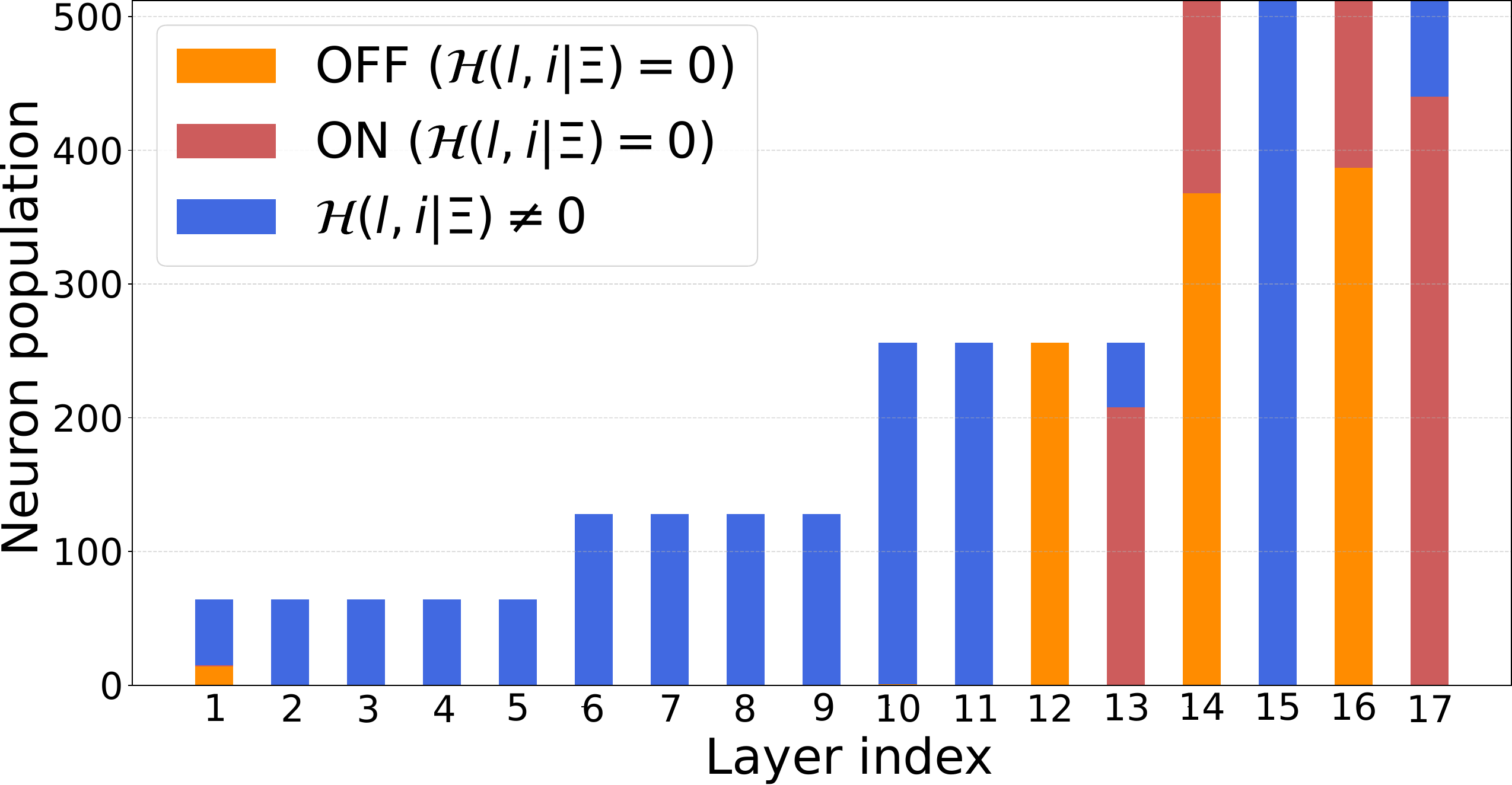}

\caption{Distributions of neuron states per layer for ResNet-18 trained on CIFAR-10 with 98.4\% of pruned parameters for EGP. In blue neurons having non-zero entropy, in orange always OFF, and in red always ON.} 
\label{ResNet18-bar}
\end{figure}

\begin{table}[t]
\centering
\caption{Performance (Top-1) and number of layers removed (Lay. rem.) for different architecture/datasets.}
\label{model_performance}
\resizebox{\columnwidth}{!}{%
\begin{tabular}{ccccccccccc}

\toprule
\textbf{Model}& \textbf{Dataset} & \bf Sparsity & \bf EGP & \bf Lay. rem. & \bf Top-1 \\ 
\midrule

\multirow{21}{*}{Resnet-18} & \multirow{10}{*}{CIFAR-10} &
                    0.0  &                             & 0/17    & 92.10\\
                    \cmidrule{3-6}
     &&             \multirow{2}{*}{50.0} &             & 0/17    & 92.56\\
     &&              &\cgr \checkmark   &\cgr 1/17    &\cgr 92.36\\
     \cmidrule{3-6}
     &&             \multirow{2}{*}{75.0} &             & 0/17    & 93.00\\
     &&              &\cgr\checkmark   &\cgr 1/17    &\cgr 92.81\\
     \cmidrule{3-6}
     &&             \multirow{2}{*}{93.8} &             & 0/17    & 93.03\\
     &&              &\cgr\checkmark   & \cgr\bf 3/17    &\cgr 92.93\\
     \cmidrule{3-6}
     &&             \multirow{2}{*}{98.4} &             & 0/17    & 92.73\\
     &&              &\cgr\checkmark   &\cgr \bf 3/17    &\cgr \bf 93.12\\
\cmidrule{2-6}
 & \multirow{10}{*}{Tiny-INet}&
                    0.0  &                             & 0/17    & \bf 41.88\\
                    \cmidrule{3-6}
     &&             \multirow{2}{*}{50.0} &             & 0/17    & 41.70\\
     &&              &\cgr\checkmark   &\cgr 2/17    &\cgr 38.96\\
     \cmidrule{3-6}
     &&             \multirow{2}{*}{75.0} &             & 0/17    & 41.24\\
     &&              &\cgr\checkmark   &\cgr 4/17    &\cgr 39.50\\
     \cmidrule{3-6}
     &&             \multirow{2}{*}{93.8} &             & 0/17    & 41.86\\
     &&              &\cgr\checkmark   &\cgr 4/17    &\cgr 39.80\\
     \cmidrule{3-6}
     &&             \multirow{2}{*}{98.4} &             & 0/17    & 36.50\\
     &&              &\cgr\checkmark   &\cgr \bf 5/17    &\cgr 37.44\\

\midrule

\multirow{21}{*}{Swin-T} & \multirow{10}{*}{CIFAR-10} &
                    0.0  &                             & 0/12    &\bf 92.11\\
                    \cmidrule{3-6}
     &&             \multirow{2}{*}{50.0} &             & 0/12    & 91.64\\
     &&              &\cgr\checkmark   &\cgr 1/12    &\cgr 92.08\\
     \cmidrule{3-6}
     &&             \multirow{2}{*}{75.0} &             & 0/12    & 89.04\\
     &&              &\cgr\checkmark   &\cgr 4/12    &\cgr \bf 92.11\\
     \cmidrule{3-6}
     &&             \multirow{2}{*}{93.8} &             & 0/12    & 84.02\\
     &&              &\cgr\checkmark   &\cgr 5/12    &\cgr 91.01\\
     \cmidrule{3-6}
     &&             \multirow{2}{*}{98.4} &             & 0/12    & 90.42\\
     &&              &\cgr\checkmark   &\cgr \bf 7/12    &\cgr 90.10\\
\cmidrule{2-6}
  & \multirow{10}{*}{Tiny-INet} &
                    0.0  &                             & 0/12    & \bf 75.38\\
                    \cmidrule{3-6}
     &&             \multirow{2}{*}{50.0} &             & 0/12    & 74.06\\
     &&              &\cgr\checkmark   &\cgr 1/12    &\cgr 71.48\\
     \cmidrule{3-6}
     &&             \multirow{2}{*}{75.0} &             & 0/12    & 72.02\\
     &&              &\cgr\checkmark   &\cgr 2/12    &\cgr 70.28\\
     \cmidrule{3-6}
     &&             \multirow{2}{*}{93.8} &             & 0/12    & 67.58\\
     &&              &\cgr\checkmark   &\cgr 3/12    &\cgr 66.58\\
     \cmidrule{3-6}
     &&             \multirow{2}{*}{98.4} &             & 0/12    & 63.46\\
     &&              &\cgr\checkmark   &\cgr \bf 6/12    &\cgr 62.30\\

\bottomrule
\end{tabular}%
}
\end{table}

\begin{table}[t]
\caption{Performance of entropy-based pruned model and reinitialized model.}
\label{Reinitialize_performance}
\resizebox{\columnwidth}{!}{%
\begin{tabular}{cc ccc}
\toprule
 \bf Model & \bf Dataset & \bf Lay. rem. & \bf Method & \bf Top-1\\
 \midrule
\multirow{4}{*}{Resnet-18} & \multirow{2}{*}{CIFAR-10} & \multirow{2}{*}{5/17} & from scratch & 91.10 \\
            &          &   &EGP &\bf 92.18\\
            \cmidrule{2-5}
 & \multirow{2}{*}{Tiny-ImageNet} & \multirow{2}{*}{4/17} & from scratch & 37.72\\
            &           &   &EGP &\bf 39.80  \\
            \midrule
\multirow{4}{*}{Swin-T} & \multirow{2}{*}{CIFAR-10} & \multirow{2}{*}{5/12} & from scratch & 61.00 \\
        &               &   &EGP &\bf 90.41\\
        \cmidrule{2-5}
  & \multirow{2}{*}{Tiny-ImageNet}  & \multirow{2}{*}{1/12} & from scratch & 0.50\\
        &               &   &EGP  &\bf 71.48\\\bottomrule
\end{tabular}%
}
\end{table}

In this section, we present our empirical results obtained on three different very common setups. We have performed our experiments on an NVIDIA GeForce RTX 2080 GPU and developed the code using PyTorch~1.13.1.
 
\textbf{Setup.} We present results for ResNet-18 and Swin-T trained on CIFAR-10 and Tiny-ImageNet (Tint-INet). While for ResNet-18 ReLU is the non-linearity for the layers, in Swin-T the non-linear activation adopted is GELU. We adopt the same baseline training strategy as in~\cite{he2022sparse}. Specifically for Swin-T, we prune just the GELU-activated layers (which are, for this model, 12). For all the sparse configurations, we set $\zeta=0.5$.\footnote{We acknowledge this choice is computationally extensive and unnecessary with optimal hyper-parameter optimization: our objective here is not to be efficient at train time but to highlight neurons with zero entropy.}

\textbf{Results.} The main results are reported in Table~\ref{model_performance}. Here, we compare a vanilla iterative magnitude pruning strategy with EGP. It appears evident that plugging EGP as a scaling factor to drive the pruning process effectively removes entire layers from the model. Evidently, with a higher sparsity, a performance drop naturally occurs; however, the gap in performance between vanilla pruning and EGP (for the same sparsity) remains consistently narrow. We provide, in Fig.~\ref{ResNet18-bar}, a visualization of the activation distributions for CIFAR-10 at 98.4\% sparsity for EGP. Interestingly, we observe that one layer only (\#12) can be removed with traditional pruning approaches, while others (like \#14), having also neurons in the ON state, would be overlooked: this shows the effectiveness of EGP and potentially opens the road to a new class of pruning algorithms.
 
\textbf{Ablation study.} We propose here an ablation study on the effectiveness of EGP. Table~\ref{Reinitialize_performance} compares training from scratch a model with some layers removed to the same following EGP: we observe that in all cases EGP exhibits better Top-1 performance. Especially for Swin-T trained on Tiny-ImageNet, this architecture/dataset combination does not provide good results as the smaller model should be also pre-trained on a large dataset like the vanilla one. This phenomenon
suggests that EGP enhances accessibility to rare, compressed states of the deep model.

%% file: sections/5_conclusion.tex
\section{Conclusion}
\label{sec:conclusion}

In this work, we have investigated the impact of unstructured pruning on the entropy of activations in DNNs. Our preliminary experiments on ReLU and GELU-activated models show that our pruning strategy, EGP, drives a consistent part of the neurons in the model to either ON or OFF regions. Unstructured pruning can provide a structured activation pattern and yes, it can reduce the model's depth! Our results provide new insights into the mechanisms behind the benefits of pruning and open avenues for further exploration into the relationship between entropy, pruning, and deep learning performance.\\
Future work can explore the applicability of our findings to other activation functions and architectures as well as the design of a regularization function explicitly enforcing low entropy to favor the model's compressibility. Exploring the impact of pruning on the dynamics of the optimization process and the interpretability of the resulting models could lead to a better understanding of the mechanisms behind the benefits of pruning. 